\documentclass{article}
\usepackage{ijcai18}

\usepackage{url}
\usepackage{times}
\usepackage{xcolor}
\usepackage{graphicx}
\usepackage{soul}
\usepackage[utf8]{inputenc}
\usepackage[small]{caption}
\usepackage{amsmath,amsfonts,amssymb,amsthm}
\usepackage{amssymb}
\usepackage{comment}

\usepackage[ruled]{algorithm}
\usepackage{algpseudocode}

\title{Counterexample-Guided Data Augmentation}

\author{
Tommaso Dreossi, 
Shromona Ghosh, 
Xiangyu Yue,\\
{\bf Kurt Keutzer,
Alberto Sangiovanni-Vincentelli, 
Sanjit A. Seshia}
\\ 
University of California, Berkeley \\
\{dreossi $\mid$ shromona.ghosh $\mid$ xyyue\}@berkeley.edu,\\
\{keutzer $\mid$ alberto $\mid$ sseshia\}@eecs.berkeley.edu
}

\newcommand{\aset}{\mathbb{A}}
\newcommand{\bset}{\mathbb{B}}

\newcommand{\sset}{\mathbb{S}}

\newcommand{\norm}[1]{\| #1 \|}

\newcommand{\va}{\mathbf{a}}

\newcommand{\vm}{\mathbf{m}}

\newcommand{\vx}{\mathbf{x}}
\newcommand{\vy}{\mathbf{y}}

\newcommand{\featsp}{\mathbb{X}}
\newcommand{\cset}{\mathbb{C}}

\newcommand{\modsp}{\mathbb{M}}

\newcommand{\trainsetsym}{\mathbb{X}}
\newcommand{\missetsym}{\cset}
\newcommand{\trainset}{\trainsetsym}
\newcommand{\misset}[1]{\missetsym^{[#1]}}
\newcommand{\tmod}[1]{f_{#1}}
\newcommand{\testset}{\mathbb{T}}
\newcommand{\ap}[2]{ap_{#1}(#2)}
\newcommand{\ar}[2]{ar_{#1}(#2)}
\newcommand{\acc}[2]{acc_{#1}(#2)}
\newcommand{\iou}[2]{IoU(#1,#2)}
\newcommand{\pre}[2]{p(#1,#2)}
\newcommand{\rec}[2]{r(#1,#2)}

\newcommand{\md}{\vm}

\newcommand{\imd}[1]{\md^{(#1)}}

\newcommand{\genf}{\gamma}

\newcommand{\dist}[2]{d(#1,#2)}

\newcounter{myctr}
\newenvironment{mylist}{\begin{list}{\arabic{myctr}.}
{\usecounter{myctr}
\setlength{\topsep}{0.3mm}\setlength{\itemsep}{0mm}
\setlength{\parsep}{0.3mm}
\setlength{\itemindent}{0mm}\setlength{\partopsep}{0mm}
\setlength{\labelwidth}{15mm}
\setlength{\leftmargin}{4mm}}}{\end{list}}

\newenvironment{myitemize}{\begin{list}{$\bullet$}
{\setlength{\topsep}{1mm}\setlength{\itemsep}{0.25mm}
\setlength{\parsep}{0.1mm}
\setlength{\itemindent}{0mm}\setlength{\partopsep}{0mm}
\setlength{\labelwidth}{15mm}
\setlength{\leftmargin}{4mm}}}{\end{list}}

\begin{document}

\maketitle

\begin{abstract}
We present a novel framework for augmenting data sets for
machine learning based on {\it counterexamples}. Counterexamples
are misclassified examples that have 
important properties for retraining and improving the model.
Key components of our framework include a \textit{counterexample generator},
which produces data items that are misclassified by the model and
{\it error tables}, a novel data
structure that stores information pertaining to misclassifications.
Error tables can be used to explain the model's
vulnerabilities and are used to efficiently generate counterexamples for augmentation.
We show the efficacy of the proposed framework by comparing it
to classical augmentation techniques on a case study of object detection in autonomous
driving based on deep neural networks.

\end{abstract}


\section{Introduction}\label{sec:introduction}

%


Models produced by machine learning algorithms, 
especially {\em deep neural networks},
are being deployed in domains where trustworthiness is a big
concern, creating the need for higher accuracy and assurance~\cite{russell2015letter,seshia-arxiv16}.
However, learning high-accuracy models using deep learning is limited by the need for large amounts of data, and, even further, by the need of labor-intensive labeling.
\emph{Data augmentation} overcomes the lack of data by inflating training sets
with label-preserving transformations, i.e., transformations which do not alter the label. Traditional data augmentation 
schemes~\cite{dataAugmentation,simard2003best,cirecsan2011high,ciregan2012multi,krizhevsky2012imagenet} involve geometric transformations which alter the geometry of the image 
(e.g., rotation, scaling, cropping or flipping); and photometric transformations which vary color channels.
The efficacy of these techniques have been demonstrated  
recently (see, e.g.,~\cite{xu2016improved,wong2016understanding}).
Traditional augmentation schemes, like the aforementioned methods, add data to the training set hoping to improve the model accuracy without taking
into account what kind of features the model has already learned.
More recently, a sophisticated data augmentation technique has been proposed~\cite{liang2017recurrent,marchesi2017megapixel} which uses
Generative Adversarial Networks~\cite{goodfellow2014generative}, a particular kind of neural network
able to generate synthetic data, to inflate training sets.
There are also augmentation techniques, such as hard negative mining~\cite{shrivastava2016training}, that inflate the training set
with targeted negative examples with the aim of reducing false positives.

In this work, we propose a new augmentation scheme, \emph{counterexample-guided data augmentation}.
The main idea is to augment the training set only with new misclassified examples rather than 
modified images coming from the original training set. The proposed augmentation scheme consists of the 
following steps: 1) Generate synthetic images that are misclassified by the model, i.e., the counterexamples;
2) Add the counterexamples to the training set; 3) Train the model on the augmented dataset.
These steps can be repeated until the desired accuracy is reached.
Note that our augmentation scheme depends on the ability to generate misclassified images.
For this reason, we developed an \emph{image generator} that cooperates with a \emph{sampler} to
produce images that are given as input to the model. 
The images are generated in a manner such
that the ground truth labels can be automatically added. 
The incorrectly classified images
constitute the augmentation set that is added to the training set.
In addition to the pictures, the image generator provides information on the misclassified
images, such as the disposition of the elements, brightness, contrast, etc. This information
can be used to find features that frequently recur in counterexamples. 
We collect information about the counterexamples in a data
structure we term as the ``\emph{error table}''.
Error tables are extremely useful to provide \emph{explanations} about
counterexamples and find recurring patterns that can lead an image to be misclassified.
The error table analysis can also be used to generate images which are 
likely to be counterexamples, and thus, efficiently build augmentation sets.

In summary, the main contributions of this work are:
\begin{myitemize}
	\item A \emph{counterexample-guided data augmentation} approach where only misclassified examples
		are iteratively added to training sets;
	\item A synthetic \emph{image generator} that renders realistic counterexamples;
	\item \emph{Error tables} that store information about counterexamples and whose analysis 
		provides explanations and facilitates the generation of counterexample images.
\end{myitemize}

We conducted experiments on
Convolutional Neural Networks (CNNs) for object detection by analyzing different
counterexample data augmentation sampling schemes and compared the 
proposed methods with classic data augmentation.
Our experiments show the benefits of using a counterexample-driven approach
against a traditional one. The improvement comes from the fact that a counterexample
augmentation set contains information that the model had not been able to learn from the training set,
a fact that was not considered by classic augmentation schemes.
In our experiments, we use synthetic data sets generated by our image generator.
This ensures that all treated data comes from the same distribution.

\subsection*{Overview}

\begin{figure}
	\centering
	\includegraphics[width=0.5\textwidth]{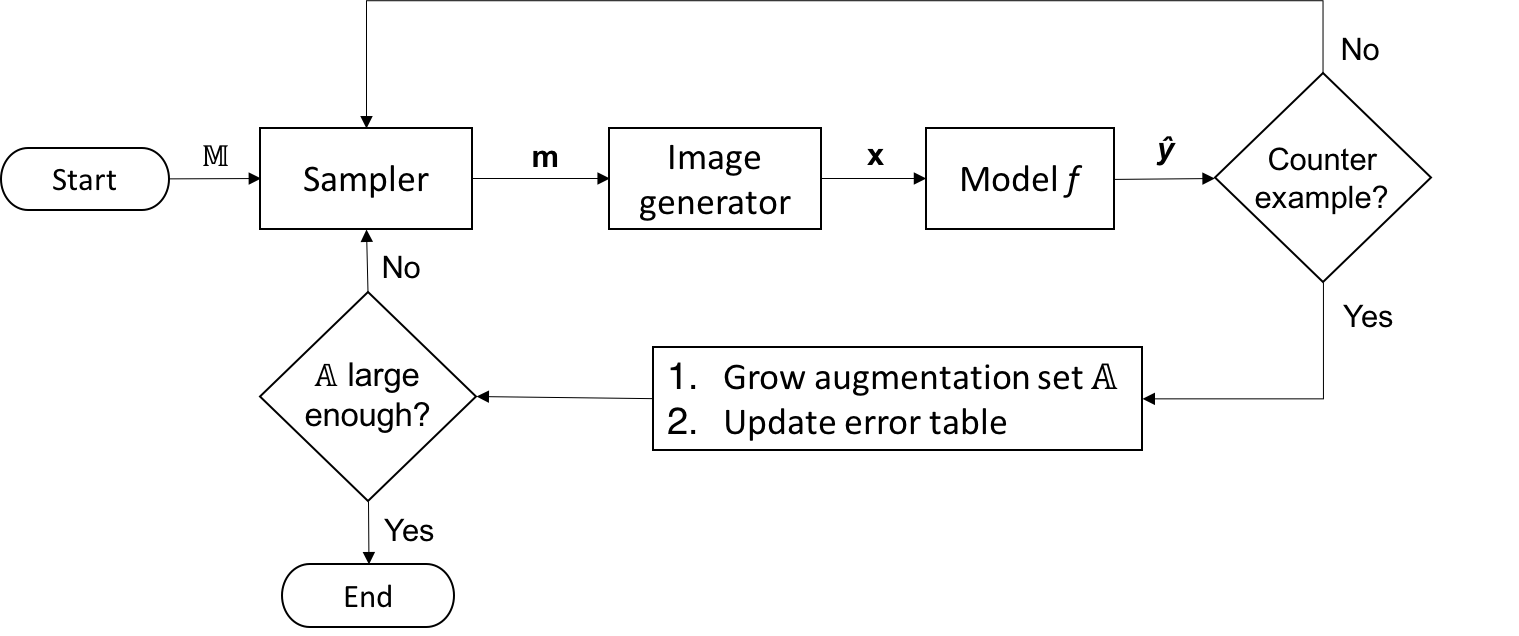}
	\caption{Counterexample-guided augmentation scheme. $\modsp$ denotes the modification space, 
	$\vm$ is a sampled modification, $\vx$ is the image generated from $\vm$, $\hat{\vy}$ is the model prediction.
	\label{fig:scheme}}
\end{figure}

Fig.~\ref{fig:scheme} summarizes the proposed counterexample-guided
augmentation scheme. The procedure takes as input
a modification space, $\modsp$, the space of possible configurations of our image generator.
The space $\modsp$ is constructed based on domain knowledge to be
a space of ``semantic modifications;'' i.e., each modification must
have a meaning in the application domain in which the machine
learning model is being used. This allows us to perform
more meaningful data augmentation than simply through
adversarial data generation
performed by perturbing an input vector (e.g., adversarially selecting
and modifying a small number of pixel values in an image).

In each loop, the sampler selects a modification, $\vm$, from $\modsp$.
The sample is determined by a sampling method that can be biased by a precomputed
\emph{error table}, a data structure
that stores information about images that are misclassified by the model.
The sampled modification is rendered into a picture $\vx$ 
by the image generator. The image $\vx$ is given as
input to the model $f$ that returns the prediction
$\hat{\vy}$. We then check whether $\vx$ is a counterexample,
i.e., the prediction $\hat{\vy}$ is wrong.
If so, we add $\vx$ to our augmentation set $\aset$
and we store $\vx$'s information (such as $\vm$, $\hat{\vy}$)
in the error table that will be used by the sampler at the next iteration.
The loop is repeated until the
augmentation set $\aset$ is large enough (or $\modsp$ has been sufficiently covered). 

This scheme returns an {\em augmentation set}, that will
be used to retrain the treated model, along with an 
{\em error table}, whose analysis
identifies common features among counterexamples and
aids the sampler to select candidate counterexamples.


The paper structure mostly follows the scheme
of Fig.~\ref{fig:scheme}:
Sec.~\ref{sec:preliminaries} introduces some notation;
Sec.~\ref{sec:image_generator}
describes the image generator used to render synthetic images; Sec.~\ref{sec:sampling} introduces some
sampling techniques that can be used to efficiently sample
the modification space; Sec.~\ref{sec:error_analysis} introduces
error tables and details how they can be used to provide 
explanations about counterexamples;
Sec.~\ref{sec:targeted_augmentation} concludes the paper by evaluating
the proposed techniques and comparing across different tunings of our
counterexample-guided augmentation scheme and the proposed
methods against classic augmentation.
The implementation of the proposed framework
and the reported experiments are
available at \url{https://github.com/dreossi/analyzeNN}.


\section{Preliminaries}\label{sec:preliminaries}

This section provides the notation used throughout this paper.

Let $\va$ be a vector, $a_i$ be its $i$-th element with
index starting at $i=1$, $a_{i:j}$ be the range of elements of $\va$ from $i$ to $j$; and $\aset$ be a set.
$\trainset$ is a set of training examples, $\vx^{(i)}$ is the $i$-th example
from a dataset and $\vy^{(i)}$ is the associated label. $f : \aset \to \bset$ is a model (or function)
$f$ with domain $\aset$ and range $\bset$.
$\hat{\vy} = f(\vx)$ is the prediction of the model $f$ for input $\vx$. In the object detection context, $\hat{\vy}$ encodes bounding 
boxes, scores, and categories predicted by $f$ for the image $\vx$.
$f_\trainset$ is the model $f$ trained on $\trainset$. 
Let $B_1$ and $B_2$ be bounding boxes encoded by $\hat{\vy}$.
The \emph{Intersection over Union} (IoU) is defined as $\iou{A_{B_1}}{A_{B_2}} = A_{B_1}\cap A_{B_2} / A_{B_1} \cup A_{B_2}$, where $A_{B_i}$ is the area of $B_i$, with $i \in \{1,2\}$.
We consider $B_{\hat{\vy}}$ to be a \emph{detection} for $B_\vy$ if $\iou{B_{\hat{\vy}}}{B_{\vy}} > 0.5$.
\emph{True positives} $tp$ is the number of correct detections;
\emph{false positives} $fp$ is the number of predicted boxes that
do not match any ground truth box; \emph{false negatives} is the 
number of ground truth boxes that are not detected.

\emph{Precision} and \emph{recall} are defined as $\pre{\hat{\vy}}{\vy} = tp / (tp + fp)$ and $\rec{\hat{\vy}}{\vy} = tp / (tp + fn)$.
In this work, we consider an input $\vx$ to be
 \emph{misclassified} if $\pre{\hat{\vy}}{\vy}$ or
$\rec{\hat{\vy}}{\vy}$ is
less than $0.75$.
Let $\testset = \{ (\vx^{(1)}, \vy^{(1)}), \dots, (\vx^{(m)}, \vy^{(m)}) \}$ be a test set with $m$ examples.
The \emph{average precision} and \emph{recall} of $f$ are defined as
	$\ap{f}{\testset} = \frac{1}{m} \sum_{i = 1}^m p(f(\vx^{(i)}),\vy^{(i)})$
	$\ar{f}{\testset} = \frac{1}{m} \sum_{i = 1}^m r(f(\vx^{(i)}),\vy^{(i)})$.
We use average precision and recall to measure the accuracy of a model, succinctly represented as
$\acc{f}{\testset} = (\ap{f}{\testset}, \ar{f}{\testset})$.


\section{Image Generator}\label{sec:image_generator}

At the core of our counterexample augmentation scheme is an 
image generator (similar to the one defined in~\cite{dreossi-nfm17,dreossi-rmlw17}) that renders realistic synthetic images of road scenarios. 
Since counterexamples are generated by the synthetic data generator, 
we have full knowledge of the ground truth labels for the generated data. 
In our case, for instance, when the image generator places a car 
in a specific position, we know exactly its location and size, hence 
the ground truth bounding box is accordingly determined.  
In this section, we describe the details of our image generator.

\subsection{Modification Space}\label{sec:mod_space}

The image generator implements a \emph{generation function}
$\genf : \modsp \to \featsp$ that maps every modification $\md \in \modsp$ 
to a feature $\genf(\md) \in \featsp$. Intuitively, a modification describes the 
configuration of an image.
For instance, a three-dimensional modification space can characterize
a car $x$ (lateral) and $z$ (away) displacement on the road and the image brightness.
A generator can be used to abstract and compactly represent a subset of a
high-dimensional image space.

\begin{figure}
	\centering
	\includegraphics[scale=0.19]{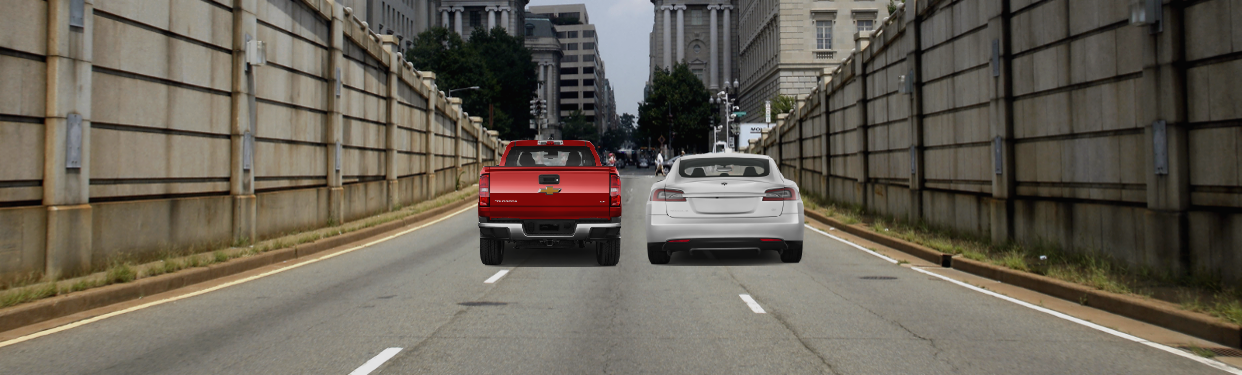}\\
	(a) {\scriptsize $\imd{1} = (53, 25, 2, 0.11, 0.98, \dots, 0.50, 0.41, \dots)$}\\[0.2cm]
	\includegraphics[scale=0.19]{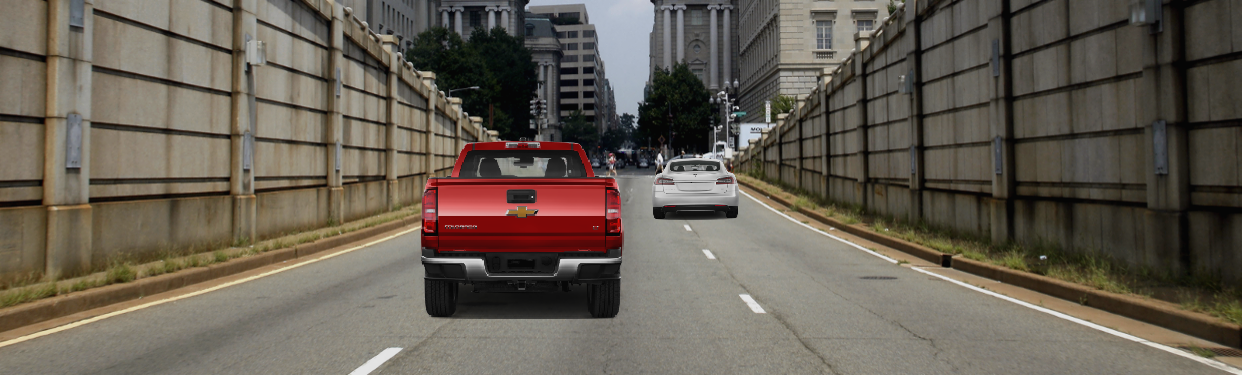}\\
	(b) {\scriptsize $\imd{2} = (53, 25, 2, 0.11, 0.98, \dots, 0.20, 0.80, \dots)$}\\[0.2cm]
	\includegraphics[scale=0.19]{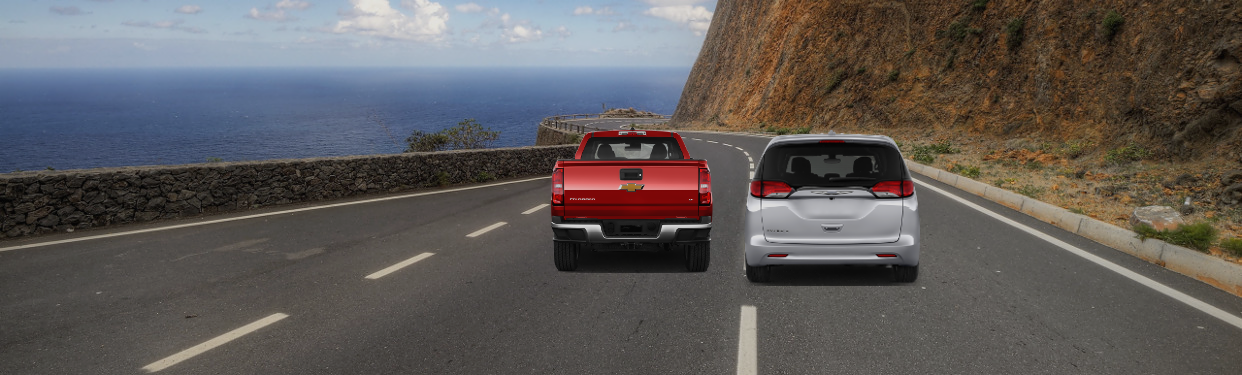}\\
	(c) {\scriptsize $\imd{3} = (13, 25, 7, 0.11, 0.98, \dots, 0.50, 0.41, \dots)$}\\[0.2cm]
	\caption{Distance over modification space
	used to measure visual diversity of concretized images. $\dist{\imd{1}}{\imd{2}} = 0.48$, $\dist{\imd{1}}{\imd{3}} = 2.0$, $\dist{\imd{2}}{\imd{3}} = 2.48$.\label{fig:diversity}}
\end{figure}

We implemented an image generator based on
a 14D modification space whose dimensions determine a road background; number of cars~(one, two or three) and their $x$ and $z$ positions
on the road; brightness, sharpness, contrast, and color of the picture.
Fig.~\ref{fig:diversity} depicts some images rendered by our image generator.

We can 
define a metric over the modification space to measure the
diversity of different pictures. Intuitively, the distance between two
configurations is large if the concretized images are visually
diverse and, conversely, it is small if the concretized images 
are similar. 

The following is an example of metric distance that can be defined 
over our 14D modification space.
Let $\imd{1}, \imd{2} \in \modsp$ be modifications. The distance is defined as:
\begin{equation}
	\dist{\imd{1}}{\imd{2}} = \sum_{i=1}^{4} \mathbf{1}_{m^{(1)}_i \neq m^{(2)}_i} +  \norm{m^{(1)}_{5:14} - m^{(2)}_{5:14}}
\end{equation}
where $\mathbf{1}_{condition}$ is $1$ if the condition is true, $0$ otherwise, and $\norm{\cdot}$ is the $L^2$ norm.
The distance counts the differences between background and car models and adds the Euclidean distance
of the points corresponding to $x$ and $z$ positions, brightness, sharpness, contrast, and color of the images.

Fig.~\ref{fig:diversity} depicts three images with their modifications $\imd{1}, \imd{2},$ and $\imd{3}$.
For brevity, captions report only the dimensions that differ among the 
images, that are background, car models and $x, z$ positions.
The distances between the modifications are $\dist{\imd{1}}{\imd{2}} = 0.48$, $\dist{\imd{1}}{\imd{3}} = 2.0$, $\dist{\imd{2}}{\imd{3}} = 2.48$.
Note how similar images, like Fig.~\ref{fig:diversity} (a) and (b) (same backgrounds and car models, slightly different car positions),
have smaller distance ($\dist{\imd{1}}{\imd{2}} = 0.48$) than diverse images, like Fig. (a) and (c); or (b) and (c) (different backgrounds, car models, and vehicle positions),
whose distances are $\dist{\imd{1}}{\imd{3}} = 2.0$ and $\dist{\imd{2}}{\imd{3}} = 2.48$.

Later on, we use this metric to
generate sets whose elements ensure a certain 
amount of diversity. (see Sec.~\ref{sec:aug_comp})

\subsection{Picture Concretization}

Once a modification is fixed, our picture generator renders
the corresponding image. The concretization is done by superimposing basic images (such as road background and vehicles)
and adjusting image parameters (such as brightness, color, or contrast) accordingly to the values specified by the modification.
Our image generator comes with a database of backgrounds and car models used as basic images. Our database
consists of 35 road scenarios (e.g., desert, forest, or freeway scenes) and 36 car models
(e.g., economy, family, or sports vehicles, from both front and rear views).
The database can be easily extended or replaced by the user.


\subsection{Annotation Tool}

\begin{figure}
	\centering
	\includegraphics[scale=0.19]{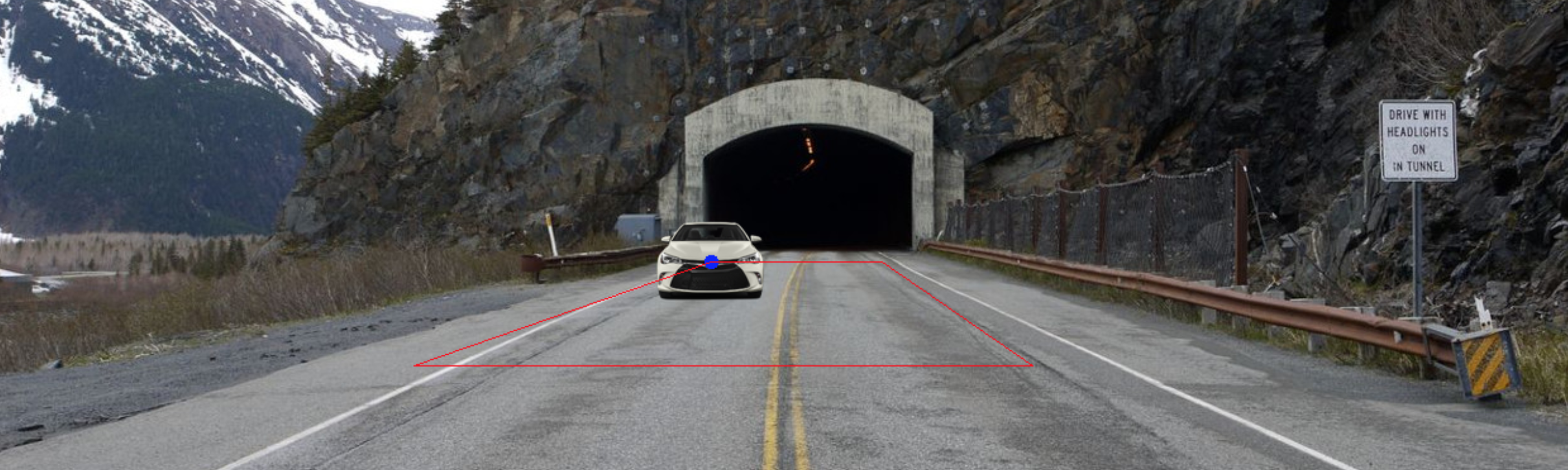}
	\caption{Annotation trapezoid. User adjusts the four corners that represent the valid sampling subspace of $x$ and $z$. The size of the car scales according to how close it is to the vanishing point.   \label{fig:annotation}}
\end{figure}

In order to render realistic images, the picture generator must 
place cars on the road and scale them accordingly.
To facilitate the conversion of a modification point describing $x$ and $z$ position 
into a proper superimposition of the car image on a road, we equipped the image generator 
with an annotation tool that can be used to specify the sampling area on a road 
and the scaling factor of a vehicle. For a particular road, the user draws a trapezoid designating the area where
the image generator is allowed to place a car. The user also specifies the scale of the car image on the trapezoid 
bases, i.e., at the closest and furthest points from the observer (see Fig.~\ref{fig:annotation}). When sampling a point at an intermediate 
position, i.e., inside the trapezoid, the tool interpolates the provided car scales and determines 
the scaling at the given point. Moreover, the image generator superimposes different
vehicles respecting the perspective of the image.
The image generator also performs several checks to ensure that the rendered cars 
are visible.

\section{Sampling Methods}\label{sec:sampling}

The goal of the sampler is to provide a good coverage of the modification space and identify samples whose concretizations lead to counterexamples.

We now briefly describe some sampling methods 
(similar to those defined in~\cite{dreossi-nfm17,dreossi-rmlw17})
that we integrated into our framework:
\begin{itemize}
\item \emph{Uniform Random Sampling}:\label{sec:rand_samp}
Uniform random sampling ensures an equal probability of sampling any possible point from $\modsp$, which guarantees a good mix of generated images for both training and testing. Although a simple and effective technique for both training as well as testing, it may not provide a good coverage of the modification space;
\item \emph{Low-Discrepancy Sampling}:\label{sec:low_disc_samp}
A \textit{low-discrepancy} (or quasi-random) sequence is a sequence of \textit{n}-tuples that fills a \textit{n}D space more uniformly than uncorrelated random points. Low-discrepancy sequences are useful to cover boxes by reducing gaps and clustering of points which ensures uniform coverage of the sample space.
\item \emph{Cross-Entropy Sampling}:
The \textit{cross-entropy} method was developed as a general Monte Carlo approach to combinatorial optimization and importance sampling. It is a iterative sampling technique, where we sample from a a given probability distribution, and update the distribution by minimizing the cross-entropy.
\end{itemize}

Some examples of low-discrepancy sequences are
the Van der Corput, Halton~\cite{halton1960efficiency}, or Sobol~\cite{sobol1976uniformly} sequences. 
In our experiments, we use the Halton~\cite{niederreiter1988low} sequence. 
There are two main advantages in having optimal coverage: first, we increase
the chances of quickly discovering counterexamples, and second, the set of counterexamples
will have high diversity; implying the concretized images will look different and thus the model will learn diverse new features.

\section{Error Tables}\label{sec:error_analysis}

Every iteration of our augmentation scheme produces a counterexample that
contains information pointing to a limitation of the learned model.
It would be desirable to extract patterns that relate counterexamples,
and use this information to efficiently generate new counterexamples.
For this reason, we define \emph{error tables} that are data structures whose columns are formed 
by important features across the generated images. The error table analysis is useful for:
\begin{enumerate}
	\item Providing \emph{explanations} about counterexamples, and
	\item Generating \emph{feedback} to sample new counterexamples.
\end{enumerate}
In the first case, by finding common patterns across counterexamples,
we provide feedback to the user
like \lq\lq\emph{The model does not detect white cars driving away from us in forest roads}\rq\rq;
in the second case, we can bias the sampler towards modifications that are
more likely to lead to counterexamples.

\subsection{Error Table Features}\label{sec:err_tab_feats}
We first provide the details of the kinds of features supported by our error tables.
We categorize features along two dimensions:
\begin{mylist}
\item {\em Explicit} vs. {\em implicit} features: Explicit features are sampled from the modification space (e.g., $x,z$ position, brightness, contrast, etc.)
	whereas implicit features are user-provided aspects of the generated image (e.g., car model, background scene, etc.).
\item {\em Ordered} vs. {\em unordered} features: some features have a domain with a well-defined total ordering (e.g., sharpness) whereas others do not have a notion of ordering (e.g., car model, identifier of background scene, etc.).
\end{mylist}
The set of implicit and explicit features are mutually exclusive. In general, implicit features are more descriptive and characterize the generated images. These are useful for providing feedback to explain the vulnerabilities of the classifier.
While implicit features are unordered, explicit features can be ordered or unordered. 
Rows of error tables are the realizations of the features for misclassification.

\begin{table}
	\caption{Example of error table proving information about counterexamples.
	First rows describes Fig.~\ref{fig:annotation}.
	Implicit unordered features: car model, environment; 
	explicit ordered features: brightness, $x,z$ car coordinates;
	explicit unordered feature: background ID.	
	\label{tab:err_tab}}
	\centering
	\resizebox{\columnwidth}{!}{
	\begin{tabular}{c | c | c | c | c | c}
		Car model & Background ID & Environment & Brightness & $x$ & $z$ \\
		\hline
		Toyota 	& $12$ & Tunnel & 0.9 & 0.2 & 0.9 \\
		BMW 	& $27$ & Forest & 1.1 & 0.4 & 0.7 \\
		Toyota 	& $11$ & Forest & 1.2 & 0.4 & 0.8 \\
	\end{tabular}
	}
\end{table}

Tab.~\ref{tab:err_tab} is an illustrative error table. The table includes car model and
environment scene (implicit unordered features), brightness, $x,z$ car coordinates
(explicit ordered features), and background ID (explicit unordered feature).
The first row of Tab.~\ref{tab:err_tab} actually refers to Fig.~\ref{fig:annotation}.
The actual error tables generated by 
our framework are larger than Tab.~\ref{tab:err_tab}. They include, for instance,
our $14$D modification space (see Sec.~\ref{sec:mod_space}) and
features like number of vehicles, vehicle orientations, dominant color 
in the background, etc.


Given an error table populated with counterexamples, we would like to analyze it to provide feedback and utilize this feedback to sample new images. 


\subsection{Feature Analysis}\label{sec:feat_analysis}
A naive analysis technique is to treat all the features equally, and search for the most commonly occurring  element in each column of the error table. However, this assumes no correlation between the features, which is often not the case. 
Instead, we develop separate analysis techniques for ordered and unordered features.
In the following we discuss how we can best capture correlations between the two sets:
\begin{itemize}
\item \emph{Ordered features}: Since these features are ordered, a meaningful analysis technique would be to find the direction in the feature space where most of the falsifying samples occur. This is very similar to model order reduction using Principal Component Analysis~(PCA~\cite{wold1987principal}). Specifically, we are interested in the first principal component, which is the singular vector corresponding to the largest singular value in the Singular Value Decomposition~(SVD~\cite{wold1987principal}) of the matrix consisting of all the samples of the ordered features.  We can use the singular vector to find how sensitive the model is with respect to the ordered features.  If the value corresponding to a feature is small in the vector, it implies that the model is not robust to changes in that feature, i.e., changes in that feature would affect the misclassification. Or alternatively, features corresponding to larger values in the singular vector, act as \lq\lq don't cares\rq\rq, i.e., by fixing all other features, the model misclassifies the image regardless the value of this feature;
\item \emph{Unordered features}: Since these features are unordered, their value holds little importance. The most meaningful information we can gather from them is the subsets of features which occurs together most often.
To correctly capture this, we must explore all possible subsets, which is a combinatorial problem. This proves to be problematic when the space of unordered features is large. One way to overcome this is by limiting the size of the maximum subset to explore. 
\end{itemize}

We conducted an experiment on a set of $500$ counterexamples.
The ordered features included $x$ and $z$ positions of each car; along with the brightness, contrast, sharpness, and color of the overall image.
The explicit features include the ordered features along with the discrete set of all possible cars and backgrounds.  The implicit features include details like color of the cars, color of the background, orientation of the cars, etc.
The PCA on the explicit ordered features revealed high values corresponding to
the $x$ position of the first car~(0.74), brightness~(0.45) and  contrast~(0.44).
We can conclude that the model is not robust to changes in these ordered features.
For the unordered features, the combination of forest road with one white car with its rear towards the camera and the other cars facing the camera, appeared $13$ times.
This provides an explanation of recurrent elements in counterexamples, specifically \lq\lq\emph{The model does not detect white cars driving away from us in forest roads}\rq\rq.

\subsection{Sampling Using Feedback}

We can utilize the feedback provided by the error table analysis to guide the sampling for subsequent training.
Note that we can only sample from the explicit features:
\begin{myitemize}
\item \emph{Feedback from Ordered Features}: The ordered features, which is a subset of the explicit features, already tell us which features need to vary more during the sampling process. For example, in the example of Sec.~\ref{sec:feat_analysis}, our sampler must prioritize sampling different $x$ positions for the first car, then brightness, and finally contrast among the other ordered features;

\item \emph{Feedback from Unordered Features}: Let $\sset_{uf} = \sset_{ef} \cup \sset_{if}$
be the subset of most occurring unordered features returned by the analysis, where $\sset_{ef}$ and $\sset_{if}$
are the mutually exclusive sets of explicit and implicit features, respectively.
The information of $\sset_{ef}$ can be directly incorporated into the sampler.
The information provided by $\sset_{if}$ require some reasoning since implicit features are not directly sampled.
However, they are associated with particular elements of the image (e.g., background or vehicle).
We can use the image generator library and error table to recognize which elements in the library the components of $\sset_{if}$ correspond to, and set the feature values accordingly.
For instance, in the example of Sec.~\ref{sec:feat_analysis}, the analysis of the unordered explicit features
revealed that the combination of bridge road with a Tesla, Mercedes, and Mazda was most often misclassified.
We used this information to generate more images with this particular combination by varying brightness and contrast.
\end{myitemize}

Sec~\ref{sec:err_exp} shows how this technique leads to a larger fraction of counterexamples that can be used for retraining.

\section{Experimental Evaluation}\label{sec:targeted_augmentation}

In this section, we show how the proposed techniques can be used
to augment training sets and improve the accuracy of the considered models.
We will experiment with different sampling methods, compare 
counterexample guided augmentation against classic augmentation, 
iterate over several augmentation cycles, and finally
show how error tables are useful tools for analyzing models.
The implementation of the proposed framework
and the reported experiments are
available at \url{https://github.com/dreossi/analyzeNN}.

In all the experiments we analyzed squeezeDet~\cite{squeezedet},
a CNN real-time object detector for autonomous driving. 
All models were trained for $65$ epochs.

The original training
and test sets $\trainset $ and $\testset$ contain $1500$ and $750$ pictures, respectively, randomly generated
by our image generator. The initial accuracy $\acc{f_\trainset}{\testset} = (0.9847,0.9843)$ 
is relatively high (see Tab.~\ref{tab:basic_aug}). However, we will be able to generate sets of counterexamples as large as $\testset$
on which the accuracy of $f_\trainset$ drops down.
The highlighted entries in the tables show the best performances.
Reported values are the averages across five different experiments.

\subsection{Augmentation Methods Comparison}\label{sec:aug_comp}

\begin{table}
	\caption{Comparison of augmentation techniques.  Precisions (top) and recalls (bottom) are reported.
	$\testset_T$ set generated with sampling method $T$;
	$f_{\trainset_T}$ model $f$ trained on $\trainset$ augmented 
	with technique $T \in \{S,R,H,C,D,M\}$; $S$: standard, $R$: uniform random, $H$: low-discrepancy Halton, $C$: cross-entropy, $D$: uniform random with distance constraint, $M$: mix of all methods.\label{tab:tech_comp}}
	\centering
	\resizebox{\columnwidth}{!}{
	\begin{tabular}{c | c c c c c}
		& $\testset_R$ & $\testset_H$ & $\testset_C$ & $\testset_D$ & $\testset_M$ \\
		\hline\\[-0.3cm]
		$f_{\trainset}$ & $\substack{0.6169 \\ 0.7429}$ & $\substack{0.6279 \\ 0.7556}$ & $\substack{0.3723\\0.4871}$ & $\substack{0.7430\\0.8373}$ & $\substack{0.6409 \\ 0.7632}$ \\[0.1cm]
		$f_{\trainset_S}$ & $\substack{0.6912 \\ 0.8080}$ & $\substack{0.6817\\ 0.7987}$ & $\substack{0.3917\\ 0.5116}$ & $\substack{0.7824\\ 0.8768}$ & $\substack{0.6994 \\ 0.8138}$ \\[0.1cm]
		$f_{\trainset_R}$ & $\substack{0.7634 \\ 0.8667}$ & $\substack{0.7515 \\ 0.8673}$ & $\substack{0.5890 \\ 0.7242 }$ & $\substack{0.8484 \\ 0.9745 }$ &  $\substack{0.7704 \\ 0.8818}$\\[0.1cm]
		$f_{\trainset_H}$ &  \colorbox{lightgray}{$\substack{0.7918 \\ 0.8673}$} & \colorbox{lightgray}{$\substack{0.7842 \\ 0.8727}$} & $\substack{0.5640 \\ 0.6693}$ & $\substack{0.8654 \\ 0.9598}$ &  \colorbox{lightgray}{$\substack{0.7980 \\ 0.8828}$}\\[0.1cm]
		$f_{\trainset_C}$ & $\substack{0.7778 \\ 0.7804}$ & $\substack{0.7632 \\ 0.7722}$ & $\substack{0.6140\\0.7013}$ &  $\substack{ 0.8673\\0.8540 }$ & $\substack{0.7843 \\ 0.7874}$\\[0.1cm]
		$f_{\trainset_D}$ & $\substack{0.7516 \\ 0.8642}$ & $\substack{0.7563 \\ 0.8724}$ & \colorbox{lightgray}{$\substack{0.6057\\0.7198}$} &  \colorbox{lightgray}{$\substack{ 0.8678\\0.9612 }$} & $\substack{0.7670 \\ 0.8815}$\\[0.1cm]
		
	\end{tabular}}
\end{table}

As the first experiment, we run the counterexample augmentation
scheme using different sampling techniques (see Sec.~\ref{sec:sampling}).
Specifically, we consider uniform random sampling, low-discrepancy
Halton sequence, cross-entropy sampling, and uniform random sampling with a diversity constraint 
on the sampled points. For the latter, we adopt the distance defined in Sec.~\ref{sec:mod_space} and
we require that the modifications of the counterexamples must be at least distant by $0.5$ from each other.

For every sampling method, we generate $1500$ counterexamples, half of which are injected
into the original training set $\trainset$ and half are used as test sets. Let $R,H,C, D$ denote 
uniform random, Halton, cross-entropy, and diversity (i.e., random with distance constraint) sampling methods.
Let $T \in \{R,H,C,D\}$ be a sampling technique. $\trainset_T$ is the
augmentation of $\trainset$, and $\testset_T$ is a test set, both
generated using $T$.  For completeness, we also defined the test set $\testset_M$ containing
an equal mix of counterexamples generated by all the $R,H,C,D$ sampling methods.

Tab.~\ref{tab:tech_comp} reports the accuracy of the models trained with various augmentation sets
evaluated on test sets of counterexamples generated with different sampling techniques.
The first row reports the accuracy of the model $f_\trainset$ trained on the original training set $\trainset$.
Note that, despite the high accuracy of
the model on the original test set ($\acc{f_\trainset}{\testset} = (0.9847,0.9843)$), we were able to generate 
several test sets from the same distribution of $\trainset$ and $\testset$ on which 
the model poorly performs.

The first augmentation that we consider is the standard one, i.e., we alter the images of $\trainset$ using
\texttt{imgaug}\footnote{imgaug: \url{https://github.com/aleju/imgaug}}, a Python library
for images augmentation. We augmented $50\%$ of the images in $\trainset$ by randomly
cropping $10-20\%$ on each side, flipping horizontally with probability $60\%$, and
applying Gaussian blur with $\sigma \in [0.0 ,3.0]$.
Standard augmentation improves the accuracies on every test set. The average precision
and recall improvements on the various test sets are $4.91\%$ and $4.46\%$, respectively (see Row 1 Tab.~\ref{tab:tech_comp}). 

Next, we augment the original training set $\trainset$ with our counterexample-guided schemes 
(uniform random, low-discrepancy Halton, cross-entropy, and random with distance constraint) and test the
retrained models on the various test sets. The average precision
and recall improvements for uniform random are $14.43\%$ and $14.56\%$, 
for low-discrepancy Halton $16.05\%$ and $14.57\%$, for cross-entropy
$16.11\%$ and $6.18\%$, and for
random with distance constraint $14.95\%$ and $14.26\%$.
First, notice the improvement in the accuracy of the original model using counterexample-guided 
augmentation methods is larger compared to the classic augmentation 
method. Second, among the proposed techniques, cross-entropy has the highest improvement in precision but low-discrepancy tends
to perform better than the other methods in average for both precision and recall.
This is due to the fact that
low-discrepancy sequences ensure more diversity on the samples 
than the other techniques, resulting in different pictures from which 
the model can learn new features or reinforce the weak ones.

The generation of a counterexample for the original 
model $f_\trainset$ takes in average for uniform random sampling
$30$s, for Halton $92$s, and for uniform random sampling with constraints 
$55$s. This shows the trade-off between time and
gain in model accuracy. The maximization of
the diversity of the augmentation
set (and the subsequent accuracy increase)
requires more iterations.

\subsection{Augmentation Loop}

For this experiment, we consider only the uniform random sampling
method and we incrementally augment the training set over several 
augmentation loops. The goal of this experiments is to understand 
whether the model overfits the counterexamples and see if 
it is possible to reach a saturation point, i.e., a model for which we 
are not able to generate counterexamples. We are also interested in
investigating the relationship between the quantity of injected counterexamples and the accuracy of the model.

Consider the $i$-th augmentation cycle. 
For every augmentation round, we generate the set of 
counterexamples by considering the model $\tmod{\trainset^{[i]}_r}$ with highest
average precision and recall. 
Given $\trainset^{[i]}$, our analysis tool
generates a set $\cset^{[i]}$ of counterexamples.
We split $\misset{i}$ in halves $\cset_\trainset^{[i]}$ and $\cset_\testset^{[i]}$.
We use $\cset_\trainset^{[i]}$ to augment the original training set $\trainset^{[i]}$ and $\cset_\testset^{[i]}$ as a test set. Specifically, the augmented training set $\trainset^{[i+1]}_{r'} = \trainset^{[i]}_{r} \cup \cset_\trainset^{[i]}$ is obtained by adding misclassified images of $\cset_\trainset^{[i]}$
to $\trainset^{[i]}$. $r, r'$ are the ratios of misclassified images to original training examples. For instance,
$|\trainset_{0.08}| = |\trainset| + 0.08*|\trainset|$, where $|\trainset|$ is the cardinality of $\trainset$.
We consider the ratios $0.08, 0.17, 0.35, 0.50$.
We evaluate every model against
every test set.

Tab.~\ref{tab:basic_aug} shows the accuracies
for three augmentation cycles. For each model, the table shows
the average precision and recall with respect to the original
test set $\testset$ and the tests sets of misclassified images.
The generation of the first loop took around
$6$ hours, the second  $14$ hours, the third 
$26$ hours. We stopped the fourth cycle after more than
$50$ hours. This shows how {\em it is increasingly hard to generate
counterexamples for models trained over several augmentations}.
This growing computational hardness of counterexample generation
with the number of cycles 
is an informal, empirical measure of increasing assurance in
the machine learning model.

\begin{table}
	\caption{Augmentation loop. For the best (highlighted) model, a test set $\cset_{\testset}^{[i]}$ and augmented training set $\trainset^{[i+1]}_r$ are generated. $r$ is the ratio of counterexamples to the original training set.\label{tab:basic_aug}}
	\centering
	\resizebox{\columnwidth}{!}{
	\begin{tabular}{c | c c c c}
			& $\testset$ & $\cset_\testset^{[1]}$ & $\cset_\testset^{[2]}$ & $\cset_\testset^{[3]}$ \\
		\hline
		$\tmod{\trainset}$ &  \colorbox{lightgray}{$\substack{0.9847\\0.9843}$} & $\substack{0.6957 \\ 0.7982}$ & & \\
		\hline
		$\tmod{\trainset^{[1]}_{0.08}}$ & $\substack{0.9842 \\ 0.9861}$ & $\substack{0.7630 \\ 0.8714}$ & & \\
		$\tmod{\trainset^{[1]}_{0.17}}$ & $\substack{0.9882 \\ 0.9905}$ &  \colorbox{lightgray}{$\substack{0.8197 \\ 0.9218}$} & $\substack{0.5922 \\ 0.8405}$ & \\
		$\tmod{\trainset^{[1]}_{0.35}}$ & $\substack{0.9843 \\ 0.9906}$ & $\substack{0.8229 \\ 0.9110}$ & & \\
		$\tmod{\trainset^{[1]}_{0.50}}$ & $\substack{0.9872 \\ 0.9912}$ & $\substack{0.7998 \\ 0.9149}$ & & \\
		\hline
		$\tmod{\trainset^{[2]}_{0.08}}$ & $\substack{0.9947 \\ 0.9955}$ & $\substack{0.7286 \\ 0.8691}$ & $\substack{0.7159 \\ 0.8612}$ & \\
		$\tmod{\trainset^{[2]}_{0.17}}$ & $\substack{0.9947 \\ 0.9954}$ & $\substack{0.7424 \\ 0.8422}$ & $\substack{0.7288 \\ 0.8628}$ & \\
		$\tmod{\trainset^{[2]}_{0.35}}$ & $\substack{0.9926 \\ 0.9958}$ & $\substack{0.7732 \\ 0.8720}$ & $\substack{0.7570 \\ 0.8762}$ & \\
		$\tmod{\trainset^{[2]}_{0.50}}$ & $\substack{0.9900 \\ 0.9942}$ & $\substack{0.8645 \\ 0.9339}$ &  \colorbox{lightgray}{$\substack{0.8223 \\ 0.9187}$} & $\substack{0.5308 \\ 0.7017}$ \\		
		\hline
		$\tmod{\trainset^{[3]}_{0.08}}$ & $\substack{0.9889 \\ 0.9972}$ & $\substack{0.7105 \\ 0.8571}$ & $\substack{0.7727 \\ 0.8987}$ & $\substack{0.7580 \\ 0.8860}$ \\
		$\tmod{\trainset^{[3]}_{0.17}}$ & $\substack{0.9965 \\ 0.9970}$ & $\substack{0.8377 \\ 0.9116}$ & $\substack{0.8098 \\ 0.9036}$&  \colorbox{lightgray}{$\substack{0.8791 \\ 0.9473}$}\\
		$\tmod{\trainset^{[3]}_{0.35}}$ & $\substack{0.9829 \\ 0.9937}$ & $\substack{0.7274 \\ 0.8060}$ & $\substack{0.8092 \\ 0.8816}$& $\substack{0.7701 \\ 0.8480}$\\
		$\tmod{\trainset^{[3]}_{0.50}}$ & $\substack{0.9905 \\ 0.9955}$ & $\substack{0.7513 \\ 0.8813}$ & $\substack{0.7891 \\ 0.9573}$& $\substack{0.7902 \\ 0.9169}$\\
	\end{tabular}}
\end{table}
 
Notice that for every cycle, our augmentation
improves the accuracy of the model with respect to the
test set. Even more interesting is the fact that the
model accuracy on the original test set does not decrease, but actually 
improves over time (at least for the chosen augmentation ratios).

\subsection{Error Table-Guided Sampling}\label{sec:err_exp}

In this last experimental evaluation, we use error tables to
analyze the counterexamples generated for $f_\trainset$ with uniform random sampling. We analyzed both the ordered  and unordered
features (see Sec.~\ref{sec:feat_analysis}).
The PCA analysis of ordered features
revealed the following relevant values:
sharpness $0.28$, contrast $0.33$, brightness $0.44$,
and $x$ position $0.77$.  This tells us that the model is more sensitive to image alterations rather than to the disposition of its elements. 
The occurrence counting of unordered features revealed that 
the top three most occurring car models in misclassifications are 
white Porsche, yellow Corvette, and light green Fiat. It is interesting
to note that all these models have uncommon designs if compared
to popular cars. The top three most recurring background scenes are a narrow bridge in a forest, an indoor parking lot, and a downtown city environment. All these scenarios are characterized by a high density
of details that lead to false positives.
Using the gathered information, we narrowed the sampler space to
the subsets of the modification space identified by the error table
analysis. The counterexample generator was able to produce
$329$ misclassification with $10$k iterations, against 
the $103$ of pure uniform random sampling, $287$ of Halton,
and $96$ of uniform random with distance constraint.

Finally, we retrained $f$ on the training set
$\trainset_E$ that includes $250$ images
generated using the error table analysis.
The obtained accuracies 
are $\acc{f_{\trainset_E}}{\testset_R} = (0.7490,0.8664)$,
$\acc{f_{\trainset_E}}{\testset_H} = (0.7582,0.8751)$,
$\acc{f_{\trainset_E}}{\testset_D} = (0.8402,0.9438)$, and 
$\acc{f_{\trainset_E}}{\testset_M} = (0.7659,0.8811)$.
Note how error table-guided sampling reaches levels of 
accuracy comparable to other counterexample-guided augmentation
schemes (see Tab.~\ref{tab:tech_comp}) but with a third of augmenting
images.

\section{Conclusion}
\label{sec:conclusion}

In this paper, we present a technique for augmenting machine
learning (ML) data sets with counterexamples. 
The counterexamples we generate are synthetically-generated
data items that are misclassified by the ML model.
Since these items are synthesized algorithmically, their
ground truth labels are also automatically generated. 
We show how error tables can be used to effectively guide the
augmentation process. Results on training deep neural networks
illustrate that our augmentation technique performs better than
standard augmentation methods for image classification.
Moreover, as we iterate the augmentation loop, it gets
computationally harder to find counterexamples.
We also show that error tables can be effective at achieving
improved accuracy with a smaller data augmentation.

We note that our proposed methodology can also be extended
to the use of counterexamples from ``system-level'' analysis
and verification, where one analyzes the correctness of an overall
system (e.g., an autonomous driving function) in the context of
a surrounding environment~\cite{dreossi-nfm17}. 
Performing data augmentation with
such ``semantic counterexamples'' is an interesting direction
for future work~\cite{dreossi-cav18}.

Our approach can be viewed as an instance of 
{\em counterexample-guided inductive synthesis} (CEGIS),
a common paradigm in program synthesis~\cite{solar-asplos06,alur-fmcad13}. In our case, the
program being synthesized is the ML model. CEGIS itself is
a special case of {\em oracle-guided inductive synthesis} (OGIS)~\cite{jha-acta17}.
For future work, it would be interesting to explore the use of
oracles other than counterexample-generating oracles to augment
the data set, and to compare our counterexample-guided data
augmentation technique to other oracle-guided data augmentation
methods.

Finally, in this work we decided to rely exclusively on 
simulated, synthesized data to make sure that the training, 
testing, and counterexample sets come from the same data source. 
It would be interesting to extend our augmentation method to
real-world data; e.g., images of road scenes collected during
driving. For this, one would need to use techniques such as
domain adaptation or transfer learning~\cite{tobin2017domain} that can adapt
the synthetically-generated data to the real world.

\section*{Acknowledgments}
This work was supported in part by
NSF grants 1545126, 1646208, and 1739816,
the DARPA BRASS program under
agreement number FA8750-16-C0043, the DARPA Assured Autonomy
program, the iCyPhy center, and Berkeley Deep Drive.
We acknowledge the support of
NVIDIA Corporation via the
donation of the Titan Xp GPU used for this research.
Hisahiro (Isaac) Ito's suggestions on cross-entropy sampling 
are gratefully acknowledged. 

%
%
%
%
%
%

\bibliographystyle{named}
\bibliography{biblio}

\begin{thebibliography}{}

\bibitem[\protect\citeauthoryear{Alur \bgroup \em et al.\egroup
  }{2013}]{alur-fmcad13}
Rajeev Alur, Rastislav Bodik, Garvit Juniwal, Milo M.~K. Martin, Mukund
  Raghothaman, Sanjit~A. Seshia, Rishabh Singh, Armando Solar-Lezama, Emina
  Torlak, and Abhishek Udupa.
\newblock Syntax-guided synthesis.
\newblock In {\em Proceedings of the IEEE International Conference on Formal
  Methods in Computer-Aided Design (FMCAD)}, pages 1--17, October 2013.

\bibitem[\protect\citeauthoryear{Ciregan \bgroup \em et al.\egroup
  }{2012}]{ciregan2012multi}
Dan Ciregan, Ueli Meier, and J{\"u}rgen Schmidhuber.
\newblock Multi-column deep neural networks for image classification.
\newblock In {\em Computer vision and pattern recognition (CVPR), 2012 IEEE
  conference on}, pages 3642--3649. IEEE, 2012.

\bibitem[\protect\citeauthoryear{Cire{\c{s}}an \bgroup \em et al.\egroup
  }{2011}]{cirecsan2011high}
Dan~C Cire{\c{s}}an, Ueli Meier, Jonathan Masci, Luca~M Gambardella, and
  J{\"u}rgen Schmidhuber.
\newblock High-performance neural networks for visual object classification.
\newblock {\em arXiv preprint arXiv:1102.0183}, 2011.

\bibitem[\protect\citeauthoryear{Dreossi \bgroup \em et al.\egroup
  }{2017a}]{dreossi-nfm17}
Tommaso Dreossi, Alexandre Donze, and Sanjit~A. Seshia.
\newblock Compositional falsification of cyber-physical systems with machine
  learning components.
\newblock In {\em Proceedings of the NASA Formal Methods Conference (NFM)},
  pages 357--372, May 2017.

\bibitem[\protect\citeauthoryear{Dreossi \bgroup \em et al.\egroup
  }{2017b}]{dreossi-rmlw17}
Tommaso Dreossi, Shromona Ghosh, Alberto~L. Sangiovanni{-}Vincentelli, and
  Sanjit~A. Seshia.
\newblock Systematic testing of convolutional neural networks for autonomous
  driving.
\newblock In {\em ICML Workshop on Reliable Machine Learning in the Wild
  (RMLW)}, 2017.
\newblock Published on Arxiv: abs/1708.03309.

\bibitem[\protect\citeauthoryear{Dreossi \bgroup \em et al.\egroup
  }{2018}]{dreossi-cav18}
Tommaso Dreossi, Somesh Jha, and Sanjit~A. Seshia.
\newblock Semantic adversarial deep learning.
\newblock In {\em 30th International Conference on Computer Aided Verification
  (CAV)}, 2018.

\bibitem[\protect\citeauthoryear{Goodfellow \bgroup \em et al.\egroup
  }{2014}]{goodfellow2014generative}
Ian Goodfellow, Jean Pouget-Abadie, Mehdi Mirza, Bing Xu, David Warde-Farley,
  Sherjil Ozair, Aaron Courville, and Yoshua Bengio.
\newblock Generative adversarial nets.
\newblock In {\em Advances in neural information processing systems}, pages
  2672--2680, 2014.

\bibitem[\protect\citeauthoryear{Halton}{1960}]{halton1960efficiency}
John~H Halton.
\newblock On the efficiency of certain quasi-random sequences of points in
  evaluating multi-dimensional integrals.
\newblock {\em Numerische Mathematik}, 2(1):84--90, 1960.

\bibitem[\protect\citeauthoryear{{Jha} and {Seshia}}{2017}]{jha-acta17}
Susmit {Jha} and Sanjit~A. {Seshia}.
\newblock {A Theory of Formal Synthesis via Inductive Learning}.
\newblock {\em Acta Informatica}, 54(7):693--726, 2017.

\bibitem[\protect\citeauthoryear{Krizhevsky \bgroup \em et al.\egroup
  }{2012}]{krizhevsky2012imagenet}
Alex Krizhevsky, Ilya Sutskever, and Geoffrey~E Hinton.
\newblock Imagenet classification with deep convolutional neural networks.
\newblock In {\em Advances in neural information processing systems}, pages
  1097--1105, 2012.

\bibitem[\protect\citeauthoryear{Liang \bgroup \em et al.\egroup
  }{2017}]{liang2017recurrent}
Xiaodan Liang, Zhiting Hu, Hao Zhang, Chuang Gan, and Eric~P Xing.
\newblock Recurrent topic-transition gan for visual paragraph generation.
\newblock {\em arXiv preprint arXiv:1703.07022}, 2017.

\bibitem[\protect\citeauthoryear{Marchesi}{2017}]{marchesi2017megapixel}
Marco Marchesi.
\newblock Megapixel size image creation using generative adversarial networks.
\newblock {\em arXiv preprint arXiv:1706.00082}, 2017.

\bibitem[\protect\citeauthoryear{Niederreiter}{1988}]{niederreiter1988low}
Harald Niederreiter.
\newblock Low-discrepancy and low-\ sequences.
\newblock {\em Journal of number theory}, 30(1):51--70, 1988.

\bibitem[\protect\citeauthoryear{Russell \bgroup \em et al.\egroup
  }{2015}]{russell2015letter}
Stuart Russell, Tom Dietterich, Eric Horvitz, Bart Selman, Francesca Rossi,
  Demis Hassabis, Shane Legg, Mustafa Suleyman, Dileep George, and Scott
  Phoenix.
\newblock Letter to the editor: Research priorities for robust and beneficial
  artificial intelligence: An open letter.
\newblock {\em AI Magazine}, 36(4), 2015.

\bibitem[\protect\citeauthoryear{Seshia \bgroup \em et al.\egroup
  }{2016}]{seshia-arxiv16}
Sanjit~A. Seshia, Dorsa Sadigh, and S.~Shankar Sastry.
\newblock {Towards Verified Artificial Intelligence}.
\newblock {\em ArXiv e-prints}, July 2016.

\bibitem[\protect\citeauthoryear{Shrivastava \bgroup \em et al.\egroup
  }{2016}]{shrivastava2016training}
Abhinav Shrivastava, Abhinav Gupta, and Ross Girshick.
\newblock Training region-based object detectors with online hard example
  mining.
\newblock In {\em Conference on Computer Vision and Pattern Recognition,
  {CVPR}}, pages 761--769, 2016.

\bibitem[\protect\citeauthoryear{Simard \bgroup \em et al.\egroup
  }{2003}]{simard2003best}
Patrice~Y Simard, David Steinkraus, John~C Platt, et~al.
\newblock Best practices for convolutional neural networks applied to visual
  document analysis.
\newblock In {\em ICDAR}, volume~3, pages 958--962, 2003.

\bibitem[\protect\citeauthoryear{Sobol}{1976}]{sobol1976uniformly}
Ilya~M Sobol.
\newblock Uniformly distributed sequences with an additional uniform property.
\newblock {\em USSR Computational Mathematics and Mathematical Physics},
  16(5):236--242, 1976.

\bibitem[\protect\citeauthoryear{Solar-Lezama \bgroup \em et al.\egroup
  }{2006}]{solar-asplos06}
Armando Solar-Lezama, Liviu Tancau, Rastislav Bod\'{\i}k, Sanjit~A. Seshia, and
  Vijay~A. Saraswat.
\newblock Combinatorial sketching for finite programs.
\newblock In {\em Proceedings of the 12th International Conference on
  Architectural Support for Programming Languages and Operating Systems
  (ASPLOS)}, pages 404--415. ACM Press, October 2006.

\bibitem[\protect\citeauthoryear{Tobin \bgroup \em et al.\egroup
  }{2017}]{tobin2017domain}
Josh Tobin, Rachel Fong, Alex Ray, Jonas Schneider, Wojciech Zaremba, and
  Pieter Abbeel.
\newblock Domain randomization for transferring deep neural networks from
  simulation to the real world.
\newblock In {\em Conference on Intelligent Robots and Systems, {IROS}}, pages
  23--30. IEEE, 2017.

\bibitem[\protect\citeauthoryear{van Dyk and Meng}{2001}]{dataAugmentation}
David~A van Dyk and Xiao-Li Meng.
\newblock The art of data augmentation.
\newblock {\em Journal of Computational and Graphical Statistics}, 10(1):1--50,
  2001.

\bibitem[\protect\citeauthoryear{Wold \bgroup \em et al.\egroup
  }{1987}]{wold1987principal}
Svante Wold, Kim Esbensen, and Paul Geladi.
\newblock Principal component analysis.
\newblock {\em Chemometrics and intelligent laboratory systems}, 2(1-3):37--52,
  1987.

\bibitem[\protect\citeauthoryear{Wong \bgroup \em et al.\egroup
  }{2016}]{wong2016understanding}
Sebastien~C Wong, Adam Gatt, Victor Stamatescu, and Mark~D McDonnell.
\newblock Understanding data augmentation for classification: when to warp?
\newblock In {\em Digital Image Computing: Techniques and Applications (DICTA),
  2016 International Conference on}, pages 1--6. IEEE, 2016.

\bibitem[\protect\citeauthoryear{Wu \bgroup \em et al.\egroup
  }{2016}]{squeezedet}
Bichen Wu, Forrest Iandola, Peter~H. Jin, and Kurt Keutzer.
\newblock Squeezedet: Unified, small, low power fully convolutional neural
  networks for real-time object detection for autonomous driving.
\newblock 2016.

\bibitem[\protect\citeauthoryear{Xu \bgroup \em et al.\egroup
  }{2016}]{xu2016improved}
Yan Xu, Ran Jia, Lili Mou, Ge~Li, Yunchuan Chen, Yangyang Lu, and Zhi Jin.
\newblock Improved relation classification by deep recurrent neural networks
  with data augmentation.
\newblock {\em arXiv preprint arXiv:1601.03651}, 2016.

\end{thebibliography}

\end{document}